\documentclass[letterpaper]{article}
\usepackage{textcomp}
\usepackage{amsmath}
\usepackage[submission]{aaai2027}  
\makeatletter
\gdef\showauthors@on{T}  
\gdef\copyright@on{}     
\makeatother
\usepackage[hyphens]{url}  
\usepackage{xcolor}
\definecolor{citeblue}{RGB}{0,82,204}
\usepackage{graphicx} 
\usepackage{natbib}  
\usepackage{caption} 
\usepackage{algorithm}
\usepackage{algorithmic}
\usepackage{newfloat}
\usepackage{multirow} 
\usepackage{amsmath}
\usepackage{amsfonts}
\usepackage{listings}
\usepackage{pifont}
\usepackage{array}
\usepackage{xcolor}
\definecolor{citeblue}{RGB}{0,82,204}
\DeclareCaptionStyle{ruled}{labelfont=normalfont,labelsep=colon,strut=off} 
\floatstyle{ruled}
\newfloat{listing}{tb}{lst}{}
\floatname{listing}{Listing}
\usepackage{booktabs}
\usepackage{amssymb}
\title{Loop-Mamba: A Loop Mamba with Degradation-Aware and Shared Memory for Old Photo Restoration}
\author {
    Runci Bai\textsuperscript{\rm 1},
    Yucheng Xin\textsuperscript{\rm 2},
    Pu Wang\textsuperscript{\rm 3},
    Yongcong Wang\textsuperscript{\rm 4},
    Chen Wu\textsuperscript{\rm 5},
    Dianjie Lu\textsuperscript{\rm 2},
    Guijuan Zhang\textsuperscript{\rm 2},
    Pengwen Dai\textsuperscript{\rm 6},
    Guangwei Gao\textsuperscript{\rm 7},
    Siyuan Yao\textsuperscript{\rm 6},
    Zhuoran Zheng\textsuperscript{\rm 8}
}
\affiliations {
    \textsuperscript{\rm 1} China Academy of Information and Communications Technology,
    \textsuperscript{\rm 2} Shandong Normal University,
    \textsuperscript{\rm 3} Shandong University,
    \textsuperscript{\rm 4} Central South University,
    \textsuperscript{\rm 5} National University of Defense Technology,
    \textsuperscript{\rm 6} Sun Yat-Sen University,
    \textsuperscript{\rm 7} Nanjing University of Science and Technology,
    \textsuperscript{\rm 8} Independent Researcher

    zhengzr@njust.edu.cn
}
\begin{document}
\maketitle
\begin{abstract}
Old photographs often suffer from multiple coupled degradations, including scratches, cracks, fading, blur, noise, and missing regions, severely degrading both visual quality and semantic content. We propose Loop-Mamba, a lightweight loop-based state-space framework that formulates old photo restoration as progressive state evolution, where a persistent restoration state is continuously propagated and refined through iterative computation. Specifically, we introduce a Semantic-Guided Degradation Estimator (SGDE) to explicitly model heterogeneous degradations by jointly predicting local degradation maps and global degradation scores, providing degradation-aware guidance for state evolution. We further develop a Shared Structural Memory Mamba (S$^2$M-Mamba), which maintains a persistent restoration state across iterations, enabling persistent state evolution through shared structural memory for robust long-range structural reconstruction. Benefiting from first-order state recursion, Loop-Mamba propagates latent restoration states through recurrent transitions instead of repeatedly stacking deep feature transformations, thereby alleviating gradient dilution while avoiding the computational overhead inherent in iterative CNN- and Transformer-based restoration frameworks. A lightweight multi-directional scanning strategy further enhances directional information aggregation and preserves structural continuity. To better evaluate restoration quality, we introduce the task-oriented Old Photo Damage Recovery Score (ODRS), which jointly measures degradation recovery and structural reconstruction fidelity. Experimental results on the public SynOld benchmark demonstrate that Loop-Mamba consistently outperforms previous state-of-the-art methods across both conventional restoration metrics and the proposed ODRS.
\end{abstract}
\section{Introduction}
\begin{figure}[t]
\centering
\includegraphics[width=1.0\columnwidth]{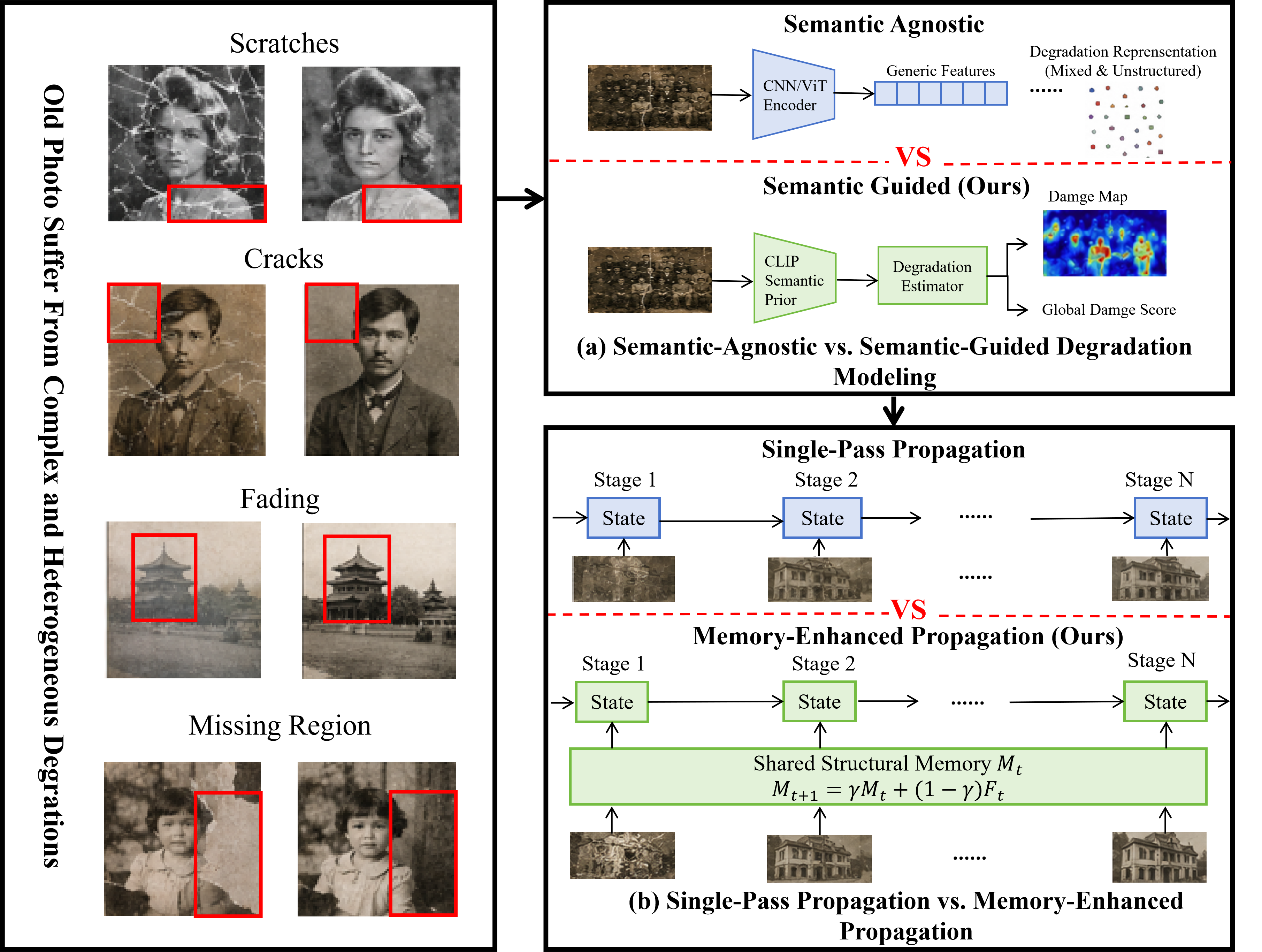}
\caption{Comparison between existing restoration paradigms and the proposed Loop-Mamba. (a) Semantic-guided degradation estimation explicitly models heterogeneous degradations by jointly predicting local degradation maps and global degradation scores, providing reliable degradation-aware guidance for progressive restoration. (b) Shared structural memory continuously propagates a persistent structural state across loop iterations, enabling progressive state evolution and structurally consistent long-range reconstruction instead of repeated feature reconstruction.}
\label{fig1}
\end{figure}
Old photographs preserve historical memories and cultural heritage. However, long-term aging, improper storage, and environmental influences often introduce multiple degradations, including scratches, cracks, fading, blur, noise, and missing regions~\cite{SSDiff}. Unlike conventional image restoration tasks focusing on a specific degradation, old photo restoration is an inherently ill-posed problem, where heterogeneous degradations severely damage structural information~\cite{Bring}. Therefore, effective restoration requires jointly modeling heterogeneous degradations while progressively reconstructing structural information.
Recent studies have shown that iterative or loop-based computation has emerged as a general paradigm for progressively refining intermediate representations across diverse learning tasks~\cite{ACT,UniversalTransformer}, demonstrating remarkable success in large language models and recurrent reasoning frameworks~\cite{Deep}. Inspired by this principle, image restoration has gradually shifted from one-shot reconstruction toward progressive refinement through recurrent computation. Representative methods such as MemNet~\cite{MemNet}, MIRNet~\cite{MIRNet}, MPRNet~\cite{MPRNet}, MAXIM~\cite{MAXIM}, and Restore, Assess, Repeat~\cite{RAR} progressively improve restoration quality via recurrent memory or multi-stage optimization. However, iterative CNN-based methods suffer from long gradient propagation paths, while Transformer-based methods incur substantial computational overhead from repeated global attention. State Space Models (SSMs) provide a more efficient alternative through linear state recursion~\cite{Mamba,Vim,VMamba}. Nevertheless, existing Mamba-based methods still rely on repeated feature refinement without explicitly modeling heterogeneous degradations or maintaining persistent restoration states.
Motivated by the above observations, we identify two key challenges for progressive loop restoration of old photographs: explicit degradation modeling and persistent state propagation. As illustrated in Fig.~1(a), we introduce a Semantic-Guided Degradation Estimator (SGDE) that explicitly models degradation severity by jointly predicting local degradation maps and global degradation scores, providing reliable degradation-aware guidance for progressive state evolution. Furthermore, instead of repeatedly reconstructing restoration states from scratch, the proposed Shared Structural Memory Mamba (S$^2$M-Mamba) continuously propagates and updates a shared structural state across restoration iterations, enabling continuous state evolution through persistent structural memory for more robust long-range structural reconstruction, as illustrated in Fig.~1(b).
Specifically, we propose Loop-Mamba, a lightweight state-space framework that formulates old photo restoration as a progressive loop state evolution process rather than repeated image reconstruction. Specifically, we first introduce a Semantic-Guided Degradation Representation to explicitly estimate both local degradation distributions and global restoration difficulty, providing degradation-aware guidance for subsequent state evolution. Building upon this representation, a shared structural memory is recursively propagated across restoration iterations, enabling long-range structural dependency modeling while preserving previously recovered content. To further enhance state propagation, we incorporate a lightweight multi-directional scanning strategy that enriches directional feature interaction during recurrent evolution. Benefiting from the linear computational complexity of State Space Models, Loop-Mamba achieves efficient progressive restoration while maintaining real-time inference. Furthermore, we introduce a task-oriented evaluation metric, termed Old Photo Damage Recovery Score (ODRS), which jointly measures degradation recovery and structural reconstruction fidelity for old photo restoration. The main \textbf{contributions} of this paper are summarized as follows:
\begin{itemize}
\item We propose Loop-Mamba, a lightweight state-space framework that reformulates old photo restoration as a progressive loop state evolution process, enabling continuous restoration through recurrent state propagation rather than repeated image reconstruction.
\item We develop a degradation-aware state evolution mechanism by integrating the Semantic-Guided Degradation Estimator (SGDE) with Shared Structural Memory Mamba (S$^2$M-Mamba) for structural restoration.
\item We introduce a multi-directional scanning strategy and the Old Photo Damage Recovery Score (ODRS) for old photo restoration.
\end{itemize}
\section{Related Work}
\subsection{Old Photo Restoration.}
Old photo restoration aims to recover photographs degraded by scratches, cracks, fading, blur, and missing regions while preserving structural consistency. Representative methods, including Bringing Old Photos Back to Life~\cite{Bring}, Bringing Old Films Back to Life~\cite{Film}, Modernizing Old Photos Using Multiple References~\cite{Modern}, and SSDiff~\cite{SSDiff}, improve restoration quality through domain translation, temporal modeling, reference guidance, or diffusion priors. However, existing methods mainly rely on external priors or powerful generative models, while explicit degradation representation and degradation-aware restoration dynamics remain largely unexplored.
\subsection{Loop-based Image Restoration.}
Loop-based image restoration progressively improves restoration quality through iterative refinement. Representative methods, including MemNet~\cite{MemNet}, MIRNet~\cite{MIRNet}, DBPN~\cite{DBPN}, MPRNet~\cite{MPRNet}, MAXIM~\cite{MAXIM}, and Restore, Assess, Repeat~\cite{RAR}, progressively reconstruct image features via recurrent memory or multi-stage refinement. However, they still rely on repeated feature reconstruction at each iteration. In contrast, our Loop-Mamba reformulates restoration as a progressive state evolution process, where a persistent hidden state is recursively propagated through first-order state recursion instead of repeatedly reconstructing image features.
\subsection{Degradation-Aware Image Restoration.}
Recent restoration methods explicitly model degradation characteristics to improve restoration quality. Representative approaches, including AirNet~\cite{AirNet}, PromptIR~\cite{PromptIR}, and DA-CLIP~\cite{DA}, learn degradation-aware representations from visual or semantic priors, demonstrating the effectiveness of degradation modeling for image restoration. However, these methods are mainly designed for generic restoration tasks and often struggle with heterogeneous degradations commonly found in old photographs.
\subsection{State Space Models for Image Restoration.}
State Space Models (SSMs) have recently emerged as an efficient alternative to Transformers owing to their linear computational complexity and long-range dependency modeling capability. Representative vision backbones, including VMamba~\cite{VMamba} and Vim~\cite{Vim}, successfully extend Mamba architectures to visual understanding, while restoration-oriented methods such as MambaIR~\cite{MambaIR} and Restormamba~\cite{Restormamba} further demonstrate their effectiveness in image restoration. Nevertheless, existing methods still lack memory-aware state propagation for robust long-range structural reconstruction.
\section{Methodology}
\begin{figure*}[t]
    \noindent\makebox[\textwidth]{\includegraphics[width=\textwidth]{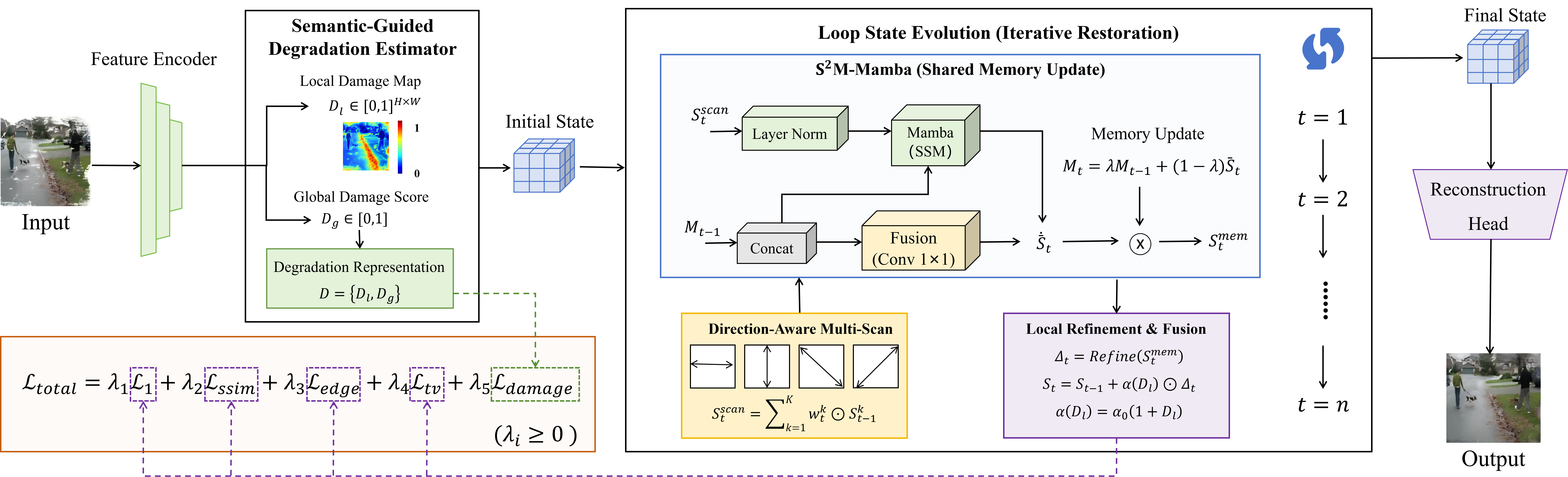}}
    \caption{Architecture of the proposed Loop-Mamba. The Semantic-Guided Degradation Estimator (SGDE) first predicts a local damage map and a global damage score to provide degradation-aware guidance. The proposed Loop State Evolution then progressively restores degraded structures through Direction-Aware Multi-Scan, Shared Structural Memory Mamba (S$^2$M-Mamba), and adaptive local refinement over multiple restoration iterations. The final restored image is reconstructed by the reconstruction head and optimized using a joint restoration objective.}
    \label{fig2}
\end{figure*}
Our proposed Loop-Mamba formulates old photo restoration as a degradation-aware loop state evolution process. An overview of the proposed framework is illustrated in Fig.~\ref{fig2}.
\subsection{Problem Formulation}
Given a degraded old photograph
$\mathbf{I}_d \in \mathbb{R}^{H\times W\times3}$,
the goal of old photo restoration is to recover its clean counterpart
$\mathbf{I}_{gt}$.
Unlike generic image restoration, old photo restoration requires modeling heterogeneous degradations while progressively reconstructing damaged structures.
Most existing methods formulate restoration as a direct image-to-image mapping,
\begin{equation}
\mathbf{I}_{r}
=
f_{\theta}
(
\mathbf{I}_{d}
),
\label{eq:traditional}
\end{equation}
\noindent
where $f_{\theta}(\cdot)$ denotes the restoration network. This formulation encodes degradation information into image features and repeatedly reconstructs states, limiting degradation modeling and long-range structural propagation.
Instead, we formulate old photo restoration as a progressive state evolution process, where a latent restoration state evolves under degradation-aware guidance,
\begin{equation}
\mathbf{S}_{t+1}
=
\mathcal{F}
(
\mathbf{S}_{t},
\mathbf{I}_{d},
\mathbf{D}
),
\label{eq:state}
\end{equation}
\noindent
where $\mathbf{D}$ denotes the degradation representation estimated from the input image and $\mathcal{F}(\cdot)$ represents the proposed state evolution operator.
After $T$ iterations, the restored image is reconstructed by
\begin{equation}
\hat{\mathbf{I}}
=
\mathcal{R}
(
\mathbf{S}_{T}
),
\label{eq:restore}
\end{equation}
\noindent
where $\mathcal{R}(\cdot)$ is the reconstruction head.
\subsection{Semantic-Guided Degradation Representation}
To explicitly characterize heterogeneous degradations, we propose a \textbf{Semantic-Guided Degradation Estimator (SGDE)} that jointly predicts a pixel-level degradation map and an image-level degradation score. The resulting degradation representation provides explicit guidance for the subsequent loop state evolution.
Given the degraded input image $\mathbf{I}_d$, the SGDE extracts degradation-aware features using a lightweight encoder,
\begin{equation}
\mathbf{F}_d
=
E_d(\mathbf{I}_d),
\label{eq:encoder}
\end{equation}
\noindent
where $E_d(\cdot)$ denotes the degradation estimation network.
The local degradation map is then predicted as
\begin{equation}
\mathbf{D}_l
=
\sigma
\left(
\phi_l(\mathbf{F}_d)
\right),
\label{eq:local}
\end{equation}
\noindent
where $\phi_l(\cdot)$ is a lightweight prediction head and $\sigma(\cdot)$ denotes the sigmoid activation. The resulting map
$\mathbf{D}_l\in[0,1]^{H\times W}$
indicates the degradation severity at each spatial location.
Meanwhile, the global degradation score is obtained by
\begin{equation}
{D_g}
=
\sigma
\left(
\phi_g
\left(
\operatorname{GAP}(\mathbf{F}_d)
\right)
\right),
\label{eq:global}
\end{equation}
\noindent
where $\operatorname{GAP}(\cdot)$ denotes global average pooling and $\phi_g(\cdot)$ is a fully connected layer.
Finally, the degradation representation is defined as
\begin{equation}
\mathbf{D}
=
\{
\mathbf{D}_l,
D_g
\},
\label{eq:damage_representation}
\end{equation}
\noindent
which is subsequently used to guide degradation-aware state evolution throughout the restoration process.
\subsection{Loop State Evolution}
Given the degradation representation $\mathbf{D}$, the restoration process is formulated as a progressive state evolution problem rather than repeated image reconstruction. Specifically, a latent restoration state is continuously propagated under degradation-aware guidance while preserving structural information through shared memory. The proposed Loop State Evolution consists of three components: degradation-modulated state update, shared structural memory evolution, and multi-directional state transition, as illustrated in Fig.~\ref{fig2}.
\subsubsection{Degradation-Guided State Update.}
The restoration state should evolve according to the spatial distribution and severity of degradations rather than following a uniform transition. To this end, the degradation representation estimated in Section~3.2 is incorporated into the state update process, allowing the restoration dynamics to adaptively focus on severely degraded regions.
Given the restoration state $\mathbf{S}_t$ and degradation representation
$\mathbf{D}=\{\mathbf{D}_l,D_g\}$,
the degradation-guided state update is formulated as
\begin{equation}
\widetilde{\mathbf{S}}_t
=
\mathbf{S}_t
+
\mathbf{D}_l
\odot
\phi(\mathbf{S}_t)
+
{D_g}
\cdot
\psi(\mathbf{S}_t),
\label{eq:guided_state}
\end{equation}
\noindent
where $\phi(\cdot)$ and $\psi(\cdot)$ denote lightweight feature transformation functions, $\odot$ represents element-wise multiplication, $\mathbf{D}_l$ provides pixel-level modulation, and $D_g$ controls the overall restoration strength.
The modulated state is then propagated through the Mamba state transition,
\begin{equation}
\mathbf{S}_{t+1}
=
\mathcal{M}
(
\widetilde{\mathbf{S}}_t
),
\label{eq:mamba_update}
\end{equation}
\noindent
where $\mathcal{M}(\cdot)$ denotes the proposed state evolution operator. In this way, degradation information is explicitly injected into the restoration dynamics, enabling adaptive state propagation according to heterogeneous degradation characteristics.
\subsubsection{Shared Structural Memory.}
Although degradation-guided state updates improve local restoration, independently evolving states gradually lose previously recovered structural information, especially under severe degradations and long restoration trajectories. To preserve global structural consistency, we introduce a \textbf{Shared Structural Memory (S$^2$M)} that continuously accumulates and propagates structural representations throughout the restoration process.
Let $\mathbf{M}_t$ denote the shared structural memory at the $t$-th iteration. After obtaining the updated state $\mathbf{S}_{t+1}$, the memory is recursively updated by
\begin{equation}
\mathbf{M}_{t+1}
=
(1-\alpha)\mathbf{M}_t
+
\alpha\mathbf{S}_{t+1},
\label{eq:memory_update}
\end{equation}
\noindent
where $\alpha\in[0,1]$ controls the memory update rate.
The updated memory is subsequently fused into the current restoration state,
\begin{equation}
\widehat{\mathbf{S}}_{t+1}
=
\mathbf{S}_{t+1}
+
\beta\mathbf{M}_{t+1},
\label{eq:memory_fusion}
\end{equation}
\noindent
where $\beta$ balances newly restored features and accumulated structural memory.
Unlike conventional iterative restoration that reconstructs restoration representations independently at each iteration, the proposed shared structural memory explicitly preserves previously recovered structures and continuously propagates them across the entire restoration process, enabling more robust long-range structural reconstruction.
\subsubsection{Multi-Directional State Propagation.}
To enhance spatial information propagation, we adopt a lightweight multi-directional scanning strategy that performs state evolution along multiple spatial directions.
Given the memory-enhanced restoration state $\widehat{\mathbf{S}}_{t+1}$, the directional state sequences are first generated by
\begin{equation}
\left\{
\mathbf{Z}^{(k)}
\right\}_{k=1}^{4}
=
\mathcal{T}
(
\widehat{\mathbf{S}}_{t+1}
),
\label{eq:scan}
\end{equation}
\noindent
where $\mathcal{T}(\cdot)$ denotes the multi-directional scanning operator. Each sequence is then propagated using the shared Mamba state evolution operator,
\begin{equation}
\mathbf{H}^{(k)}
=
\mathcal{M}
(
\mathbf{Z}^{(k)}
),
\qquad
k=1,\ldots,4.
\label{eq:direction}
\end{equation}
Finally, the propagated states are aggregated as
\begin{equation}
\mathbf{S}_{t+1}^{*}
=
\mathcal{A}
(
\mathbf{H}^{(1)},
\mathbf{H}^{(2)},
\mathbf{H}^{(3)},
\mathbf{H}^{(4)}
),
\label{eq:fusion}
\end{equation}
\noindent
where $A(\cdot)$ denotes the aggregation operator. Multi-directional propagation enriches long-range interactions with negligible computational overhead.
\subsubsection{Theoretical Analysis of State Evolution.}
Unlike conventional iterative restoration frameworks that repeatedly reconstruct feature representations, the proposed Loop-Mamba formulates restoration as a first-order state evolution process. Specifically, the recurrent structural state satisfies
\begin{equation}
\mathbf{S}_{t+1}= \mathbf{A}\mathbf{S}_{t}+\mathbf{B}\mathbf{X}_{t},
\label{eq:ssm_theory}
\end{equation}
\noindent
where $\mathbf{S}_{t}$ denotes the shared structural state, $\mathbf{X}_{t}$ is the degradation-aware feature, and $\mathbf{A}$ and $\mathbf{B}$ are the state transition matrices. Expanding Eq.~(\ref{eq:ssm_theory}) over $T$ iterations yields
\begin{equation}
\mathbf{S}_{T}
=
\mathbf{A}^{T}\mathbf{S}_{0}
+
\sum_{i=0}^{T-1}
\mathbf{A}^{T-1-i}\mathbf{B}\mathbf{X}_{i}.
\label{eq:ssm_expand}
\end{equation}
Consequently, the gradient propagated through the recurrent state satisfies
\begin{equation}
\frac{\partial \mathcal{L}}
{\partial \mathbf{S}_{t}}
=
\mathbf{A}^{\top}
\frac{\partial \mathcal{L}}
{\partial \mathbf{S}_{t+1}}.
\label{eq:grad}
\end{equation}
Compared with iterative CNN- and Transformer-based restoration, whose gradients are propagated through repeatedly stacked feature transformations, Eq.~(\ref{eq:grad}) shows that Loop-Mamba only performs first-order state recursion. Therefore, the gradient propagation path remains constant across restoration iterations, avoiding progressive gradient dilution caused by repeatedly increasing network depth. Moreover, since state evolution only requires linear state transition rather than repeatedly computing global attention, the computational complexity grows linearly with the sequence length, making Loop-Mamba naturally suitable for efficient progressive restoration.
\subsection{Optimization Objective}
Loop-Mamba is optimized in an end-to-end manner using a joint reconstruction and degradation supervision objective,
\begin{equation}
\mathcal{L}
=
\mathcal{L}_{rec}
+
\lambda_d
\mathcal{L}_{deg},
\label{eq:total_loss}
\end{equation}
\noindent
where $\lambda_d$ balances the two objectives. The reconstruction loss is defined as
\begin{equation}
\mathcal{L}_{rec}
=
\lambda_1\mathcal{L}_{L1}
+
\lambda_2\mathcal{L}_{SSIM}
+
\lambda_3\mathcal{L}_{edge}
+
\lambda_4\mathcal{L}_{TV},
\label{eq:rec_loss}
\end{equation}
\noindent
which jointly preserves pixel fidelity, structural consistency, edge continuity, and smoothness. The degradation supervision is
\begin{equation}
\mathcal{L}_{deg}
=
(D_g-\hat D_g)^2,
\end{equation}
\noindent
where $\hat{\mathbf{D}}_g$ and $\mathbf{D}_g$ denote the predicted and ground-truth global degradation scores, respectively.
\section{Experiments}
\subsection{Experimental Settings}
\subsubsection{Dataset.}
Experiments are conducted on the public SynOld benchmark (700 image pairs), following the official split of 500 training pairs (50 for validation) and 200 testing pairs. In addition, we evaluate on the Stockfilm dataset~\cite{stockfilm2026}, a real-world dataset comprising 100 clips scanned from 8mm and Super 8 home-movie reels.

\subsubsection{Metrics.}
We evaluate restoration quality using four widely adopted full-reference image quality metrics, including PSNR, SSIM~\cite{SSIM}, LPIPS~\cite{LPIPS}, and FSIM~\cite{FSIM}. In addition, considering the increasing emphasis on perceptual quality in recent Diffusion-based restoration methods, we further compare Loop-Mamba with Diffusion-based baselines using three no-reference image quality assessment (NR-IQA) metrics, namely BRISQUE~\cite{BRISQUE}, TOPIQ~\cite{TOPIQ}, and MAN-IQA~\cite{MANIQA}. We also evaluate these three NR-IQA metrics on the real-world Stockfilm dataset~\cite{stockfilm2026} to assess restoration quality under real-world degradations. To evaluate computational efficiency, we additionally report the inference speed measured by FPS, together with the computational complexity measured by FLOPs and MACs.
Although conventional restoration metrics effectively evaluate global image fidelity, they cannot faithfully measure restoration quality in severely degraded regions or quantify structural reconstruction. Therefore, we further introduce the \textbf{Old Photo Damage Recovery Score (ODRS)}, which jointly measures degradation recovery and structural reconstruction.
Specifically, the Damage Recovery Score (DRS) is defined as
\begin{equation}
\mathrm{DRS}
=
1-
\frac{
\sum_i
\mathbf{W}_i
\left|
\hat{\mathbf{I}}_i-\mathbf{I}_i
\right|
}
{
\sum_i
\mathbf{W}_i
+\varepsilon
},
\label{eq:drs}
\end{equation}
\noindent
where the degradation weight is computed by
\begin{equation}
\mathbf{W}=\mathbf{D}^{\gamma},
\label{eq:damage_weight}
\end{equation}
\noindent
with $\mathbf{D}$ denoting the degradation map estimated from the difference between degraded and clean images.
To evaluate structural reconstruction, we compute the Structural Recovery Score (SRS) using the cosine similarity between CLIP image embeddings,
\begin{equation}
\mathrm{SRS}
=
\cos
\left(
f(\hat{\mathbf{I}}),
f(\mathbf{I})
\right),
\label{eq:srs}
\end{equation}
\noindent
where $f(\cdot)$ denotes the CLIP image encoder.
Finally, ODRS is defined as
\begin{equation}
\mathrm{ODRS}
=
w_d
\mathrm{DRS}
+
w_s
\mathrm{SRS},
\label{eq:odrs}
\end{equation}
\noindent
where $w_d$ and $w_s$ are adaptive weights determined according to the statistical variation of DRS and SRS.
\subsubsection{Implementation Details.}
The proposed Loop-Mamba is implemented in PyTorch. Unless otherwise specified, all experiments are conducted on an NVIDIA RTX 4060 GPU, while RAR is evaluated on an NVIDIA RTX PRO 6000 (96\,GB) GPU due to its substantially higher computational and memory requirements. The network is optimized using the AdamW optimizer with a batch size of 4 for 400 epochs. The initial learning rate is set to $2\times10^{-4}$ with a weight decay of $1\times10^{-4}$, and is gradually decayed using a cosine annealing schedule after a 10-epoch linear warm-up. During both training and testing, all images are resized to $128\times128$. Standard data augmentation, including random horizontal flipping, random rotation, and color jittering, is applied during training. Unless otherwise specified, all restoration quality metrics are evaluated on $128\times128$ images, while efficiency metrics (FPS, FLOPs, and MACs) are benchmarked using $1920\times1080$ inputs.
\subsection{Comparison with State-of-the-Art Methods}
\subsubsection{Quantitative results.}
\begin{table*}[t]
\centering
\small
\setlength{\tabcolsep}{6pt}
\resizebox{\textwidth}{!}{
\begin{tabular}{ll>{\centering\arraybackslash}p{1.1cm}cccccccccc}
\toprule
\textbf{Category}
&
\textbf{Method}
&
\textbf{Venue \& Year}
&
\multicolumn{4}{c}{\textbf{Global Quality Metrics}}
&
\multicolumn{3}{c}{\textbf{Old Photo Damage Recovery Score}}
&
\multicolumn{3}{c}{\textbf{Efficiency Metrics ($1080P$)}}
\\
\cmidrule(lr){4-7}
\cmidrule(lr){8-10}
\cmidrule(lr){11-13}
&
&
&
\textbf{PSNR$\uparrow$}
&
\textbf{SSIM$\uparrow$}
&
\textbf{LPIPS$\downarrow$}
&
\textbf{FSIM$\uparrow$}
&
\textbf{ODRS$\uparrow$}
&
\textbf{DRS$\uparrow$}
&
\textbf{SRS$\uparrow$}
&
\textbf{FPS$\uparrow$}
&
\textbf{FLOPs(G)$\downarrow$}
&
\textbf{MACs(G)$\downarrow$}
\\
\midrule
\multirow{4}{*}{CNN-based}
& MemNet
& ICCV'17
&29.59
&0.89
&0.09
&0.9432
&0.9336
&0.9237
&0.9506
&0.07
&11990
&5995
\\
& MPRNet
& CVPR'21
&29.52
&0.89
&0.09
&0.9423
&0.9330
&0.9231
&0.9542
&1.27
&34719
&17360
\\
& NAFNet
& ECCV'22
&27.25
&0.87
&0.09
&0.9998
&0.8982
&0.9259
&0.8840
&0.55
&1995
&997
\\
& ClearAIR
& AAAI'26
&28.80
&0.81
&0.09
&0.9312
&0.8249
&0.7890
&0.9283
&--
&521
&261
\\
\midrule
\multirow{4}{*}{Transformer-based}
& SwinIR
& ICCV'21
&22.88
&0.64
&0.28
&0.9991
&0.7699
&0.8555
&0.7406
&0.78
&686
&343
\\
& Restormer
& CVPR'22
&30.09
&0.90
&0.10
&0.9485
&0.9381
&0.9568
&0.9222
&0.13
&4568
&2285
\\
& HogFormer
& AAAI'26
&25.22
&0.80
&0.25
&0.7761
&0.6752
&0.5616
&0.9077
&--
&--
&--
\\
& Restore, Assess, Repeat
& CVPR'26
&23.56
&0.76
&0.12
&0.8858
&0.8600
&0.8294
&0.8792
&0.24
&143688
&71844
\\
\midrule
\multirow{2}{*}{Diffusion-based}
& DiffBIR
& CVPR'24
&20.08
&0.47
&0.37
&0.9991
&0.8042
&0.8593
&0.7573
&1.13
&686
&343
\\
& AMFDiff
& SPIC'26
&25.32
&0.86
&0.40
&0.5809
&0.7550
&0.9270
&0.6734
&--
&1250
&625
\\
\midrule
\multirow{4}{*}{Mamba-based}
& MambaIR
& ECCV'24
&28.08
&0.90
&0.09
&0.9998
&0.8788
&0.9038
&0.8541
&0.07
&3694
&1847
\\
& MambaIRv2
& CVPR'25
&23.33
&0.83
&0.11
&0.9994
&0.8744
&0.9043
&0.8487
&0.24
&4487
&2244
\\
& MambaCS
& CVPR'26
&22.60
&0.67
&0.45
&0.7238
&0.7019
&0.8461
&0.5211
&--
&--
&--
\\
& \textbf{Loop-Mamba (Ours)}
& \textbf{-}
&
\textbf{30.13}
&
\textbf{0.93}
&
\textbf{0.09}
&
\textbf{0.9999}
&
\textbf{0.9495}
&
\textbf{0.9743}
&
\textbf{0.9419}
&
\textbf{1.86}
&
\textbf{119}
&
\textbf{59}
\\
\bottomrule
\end{tabular}
}
\caption{
Quantitative comparison with state-of-the-art old photo restoration methods on the SynOld benchmark.
$\uparrow$ and $\downarrow$ denote that higher and lower values indicate better performance, respectively.
``--'' indicates that the corresponding results are unavailable because the official implementations could not be practically benchmarked at this resolution due to excessive computational or memory requirements.
}
\label{tab:comparison}
\end{table*}
\begin{table}[t]
\centering
\renewcommand{\arraystretch}{0.85}
\setlength{\tabcolsep}{3pt}
\begin{tabular}{lccc}
\toprule
\textbf{Metric}
&
\textbf{Loop-Mamba}
&
\textbf{AMFDiff}
&
\textbf{DiffBIR}
\\
\midrule
BRISQUE$\downarrow$
&
\textbf{16.35}
&
57.67
&
16.95
\\
TOPIQ$\uparrow$
&
\textbf{0.52}
&
0.24
&
0.21
\\
MAN-IQA$\uparrow$
&
0.15
&
\textbf{0.17}
&
0.15
\\
\bottomrule
\end{tabular}
\renewcommand{\arraystretch}{1.0}
\caption{Comparison of no-reference image quality assessment (NR-IQA) metrics on the SynOld benchmark.}
\label{tab:nr_iqa}
\end{table}
Table~\ref{tab:comparison} compares Loop-Mamba with representative loop-based, CNN-based, Transformer-based, Diffusion-based, and Mamba-based restoration methods, including MemNet~\cite{MemNet}, MPRNet~\cite{MPRNet}, RAR~\cite{RAR}, NAFNet~\cite{NAFNet}, ClearAIR~\cite{ClearAIR}, SwinIR~\cite{SwinIR}, Restormer~\cite{Restormer}, HogFormer~\cite{HogFormer}, DiffBIR~\cite{DiffBIR}, AMFDiff~\cite{AMFDiff}, MambaIR~\cite{MambaIR}, MambaIRv2~\cite{MambaIRv2}, and MambaCS~\cite{MambaCS}. Loop-Mamba achieves the best PSNR, SSIM, FSIM, and ODRS while requiring substantially fewer FLOPs and MACs and achieving the highest inference speed on 1080P images, demonstrating an excellent quality-efficiency trade-off. Table~\ref{tab:nr_iqa} further compares Loop-Mamba with two recent Diffusion-based methods using no-reference image quality assessment metrics. Loop-Mamba achieves the best BRISQUE and TOPIQ scores while remaining competitive on MAN-IQA, indicating strong perceptual quality.
\subsubsection{Qualitative results.}
\begin{figure*}[t]
    \noindent\makebox[\textwidth]{\includegraphics[width=\textwidth]{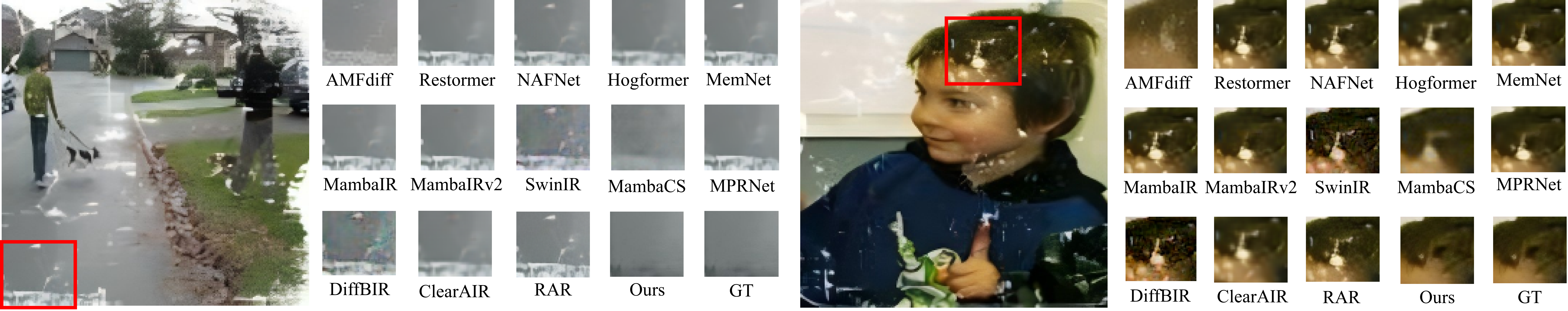}}
    \caption{Qualitative comparison with state-of-the-art methods on the SynOld benchmark.}
    \label{fig3}
\end{figure*}
Figure~\ref{fig3} presents visual comparisons on representative old photographs with diverse degradation patterns. Existing methods often leave severe degradations or introduce structural distortions. In contrast, Loop-Mamba restores cleaner structures and finer textures, producing more visually robust results.

\subsection{Ablation Study}
To comprehensively evaluate the proposed design, we conduct ablation studies on the SynOld benchmark, including the effectiveness of the proposed modules (SGDE, S$^2$M, and DMS), as well as the influence of the state dimension $d$ and the restoration cycle number $t$.
\subsubsection{Effectiveness of the Proposed Components.}
\begin{table*}[t]
\centering
\small
\setlength{\tabcolsep}{3pt}
\begin{tabular*}{\textwidth}{@{\extracolsep{\fill}} c c c c cccc ccc ccc}
\toprule
\textbf{Variant}
&
\multicolumn{3}{c}{\textbf{Proposed Modules}}
&
\multicolumn{4}{c}{\textbf{Global Quality Metrics}}
&
\multicolumn{3}{c}{\textbf{Old Photo Damage Recovery Score}}
&
\multicolumn{3}{c}{\textbf{Efficiency Metrics ($1080P$)}}\\
\cmidrule(lr){2-4}
\cmidrule(lr){5-8}
\cmidrule(lr){9-11}
\cmidrule(lr){12-14}
&
\textbf{SGDE}
&
\textbf{S$^2$M}
&
\textbf{DMS}
&
\textbf{PSNR}$\uparrow$
&
\textbf{SSIM}$\uparrow$
&
\textbf{LPIPS}$\downarrow$
&
\textbf{FSIM}$\uparrow$
&
\textbf{ODRS}$\uparrow$
&
\textbf{DRS}$\uparrow$
&
\textbf{SRS}$\uparrow$
&
\textbf{FPS}$\uparrow$
&
\textbf{FLOPs(G)}$\downarrow$
&
\textbf{MACs(G)}$\downarrow$
\\
\midrule
\textbf{V1}
&
&
&
&
29.35
&
0.91
&
0.11
&
0.9206
&
0.8432
&
0.9338
&
0.7972
&
4.55
&
14
&
7
\\
\textbf{V2}
&
\ding{51}
&
&
&
29.31
&
0.91
&
0.11
&
0.9203
&
0.9423
&
0.9741
&
0.9338
&
4.07
&
17
&
9
\\
\textbf{V3}
&
&
\ding{51}
&
&
28.47
&
0.90
&
0.11
&
0.9115
&
0.8191
&
0.7671
&
0.9283
&
3.65
&
17
&
9
\\
\textbf{V4}
&
&
&
\ding{51}
&
29.36
&
0.92
&
0.10
&
0.9227
&
0.8546
&
0.8119
&
0.9348
&
3.80
&
26
&
13
\\
\textbf{V5}
&
\ding{51}
&
\ding{51}
&
&
29.47
&
0.92
&
0.10
&
0.9246
&
0.9446
&
0.9757
&
0.9360
&
3.20
&
21
&
10
\\
\textbf{V6}
&
\ding{51}
&
&
\ding{51}
&
29.42
&
0.92
&
0.10
&
0.9230
&
0.9428
&
0.9733
&
0.9346
&
3.47
&
30
&
15
\\
\textbf{V7}
&
&
\ding{51}
&
\ding{51}
&
29.88
&
0.92
&
0.10
&
0.9282
&
0.8285
&
0.9419
&
0.8652
&
1.82
&
127
&
63
\\
\textbf{Full}
&
\ding{51}
&
\ding{51}
&
\ding{51}
&
\textbf{30.13}
&
\textbf{0.93}
&
\textbf{0.09}
&
\textbf{0.9999}
&
\textbf{0.9495}
&
\textbf{0.9743}
&
\textbf{0.9419}
&
\textbf{1.86}
&
\textbf{119}
&
\textbf{59}
\\
\bottomrule
\end{tabular*}
\caption{
Ablation study of the proposed Loop-Mamba on the SynOld benchmark.  $\uparrow$ and $\downarrow$ indicate that higher and lower values represent better performance, respectively.
}
\label{tab:ablation}
\end{table*}
\begin{figure}[t]
\centering
\includegraphics[width=1.0\columnwidth]{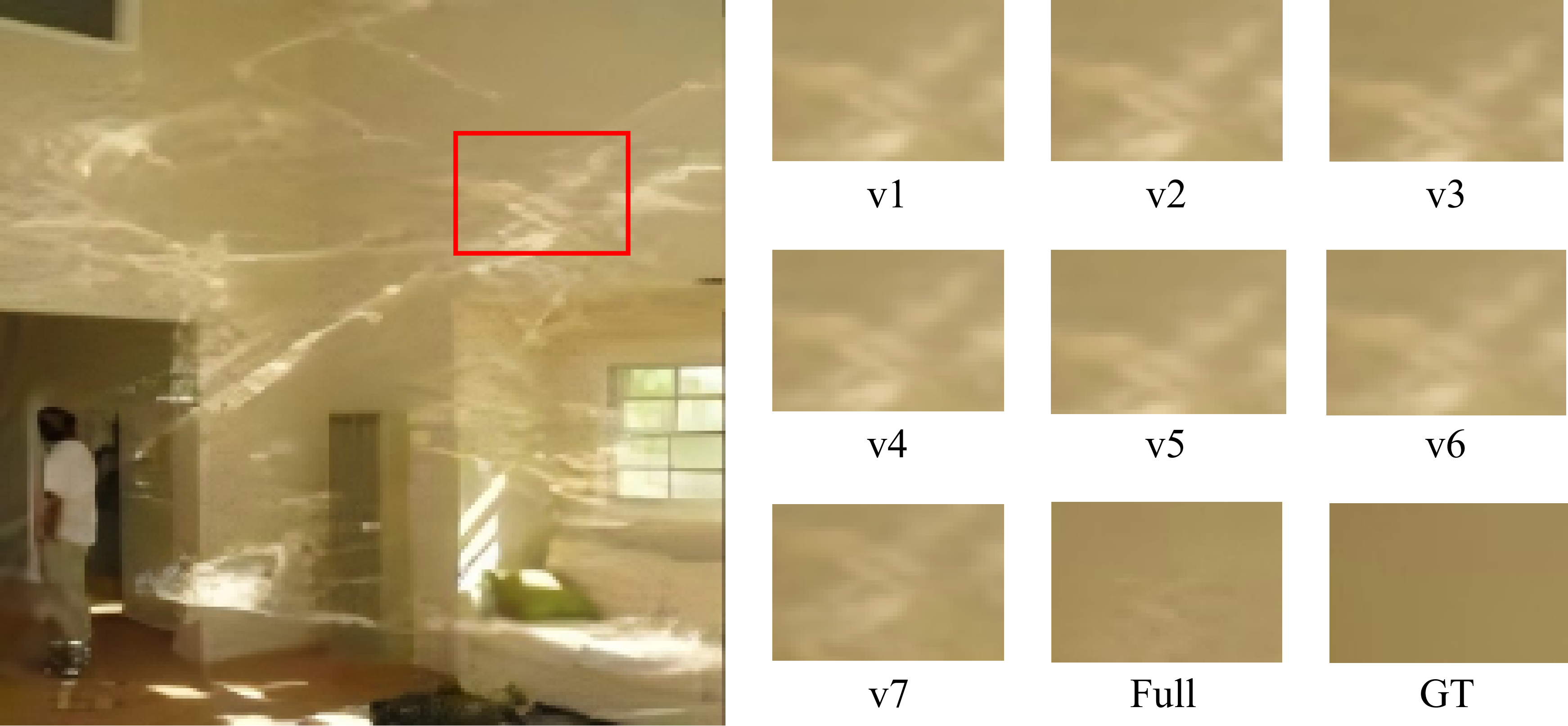}
\caption{Visual Comparison of Different Model Variants.}
\label{fig4}
\end{figure}
Table~\ref{tab:ablation} reports the contribution of each component. SGDE (V2) improves DRS and ODRS through explicit degradation modeling, while S$^2$M (V3) and DMS (V4) enhance structural consistency and local texture reconstruction, respectively. Combining all modules consistently improves performance, as further illustrated by the progressively recovered structures and textures in Fig.~\ref{fig4}.
\subsubsection{Effectiveness of State Dimension.}
\begin{table}[t]
\centering
\footnotesize
\renewcommand{\arraystretch}{0.85}
\setlength{\tabcolsep}{3pt}
\begin{tabular}{cccccc}
\toprule
\textbf{State Dim $d$}
&
\textbf{PSNR}$\uparrow$
&
\textbf{SSIM}$\uparrow$
&
\textbf{LPIPS}$\downarrow$
&
\textbf{ODRS}$\uparrow$
&
\textbf{FPS}$\uparrow$
\\
\midrule
16
&
30.13
&
0.93
&
0.0922
&
0.9495
&
\textbf{1.86}
\\
32
&
30.53
&
0.93
&
0.0852
&
0.9559
&
1.13
\\
48
&
30.98
&
0.94
&
0.0797
&
0.9596
&
0.21
\\
64
&
30.73
&
0.94
&
0.0764
&
0.9611
&
0.09
\\
\bottomrule
\end{tabular}
\renewcommand{\arraystretch}{1.0}
\caption{Ablation study on the state dimension.}
\label{tab:dim_ablation}
\end{table}
We further investigate the influence of the state dimension $d$ in Table~\ref{tab:dim_ablation}. Increasing $d$ yields diminishing performance gains at substantially higher computational cost, indicating that a moderate state dimension is already sufficient to capture most restoration information. Since larger $d$ only marginally improves PSNR and ODRS while significantly reducing inference speed, we adopt $d=16$ as the default configuration for the best quality-efficiency trade-off.
\subsubsection{Effectiveness of Restoration Cycle
Number.}
\begin{table}[t]
\centering
\footnotesize
\renewcommand{\arraystretch}{0.85}
\setlength{\tabcolsep}{3pt}
\begin{tabular}{cccccc}
\toprule
\textbf{Cycle $t$}
&
\textbf{PSNR}$\uparrow$
&
\textbf{SSIM}$\uparrow$
&
\textbf{LPIPS}$\downarrow$
&
\textbf{ODRS}$\uparrow$
&
\textbf{FPS}$\uparrow$
\\
\midrule
1
&
29.84
&
0.92
&
0.0934
&
0.9497
&
4.78
\\
2
&
29.58
&
0.92
&
0.0949
&
0.9499
&
2.67
\\
3
&
\textbf{30.13}
&
\textbf{0.93}
&
\textbf{0.0922}
&
0.9495
&
1.86
\\
4
&
29.70
&
0.93
&
0.0934
&
0.9484
&
1.42
\\
5
&
29.88
&
0.92
&
0.0947
&
0.9466
&
1.15
\\
\bottomrule
\end{tabular}
\renewcommand{\arraystretch}{1.0}
\caption{Ablation study on the number of restoration cycles.}
\label{tab:cycle_ablation}
\end{table}
We further investigate the influence of the restoration cycle number $t$ in Table~\ref{tab:cycle_ablation}. Restoration performance first improves and then gradually declines as $t$ increases, suggesting that early state evolution effectively accumulates restoration information while excessive iterations lead to slight over-refinement. Therefore, we adopt $t=3$ as the default configuration for the best quality-efficiency trade-off.

\subsection{Real-World Old Photo Restoration}

To comprehensively evaluate the practical applicability of our proposed Loop‑Mamba, we conduct experiments on real-world old photo restoration tasks. We compare Loop‑Mamba against representative methods from four mainstream architectures: CNN‑based (NAFNet, ClearAIR), Transformer‑based (Restormer, HogFormer), Diffusion‑based (DiffBIR, AMFDiff), and Mamba‑based (MambaIR, MambaIRv2). For quantitative assessment, we adopt three widely used no‑reference image quality metrics—BRISQUE, TOPIQ, and MAN‑IQA. We evaluate these methods on the Stockfilm dataset~\cite{stockfilm2026}, which comprises 100 clips scanned from 8mm and Super 8 home-movie reels. The results are summarized in Table.~\ref{tab:quality}, where our Loop‑Mamba achieves the best BRISQUE score and competitive TOPIQ and MAN‑IQA scores, demonstrating its effectiveness in perceptual quality enhancement.

\begin{figure}[t]
\centering
\includegraphics[width=1.0\columnwidth]{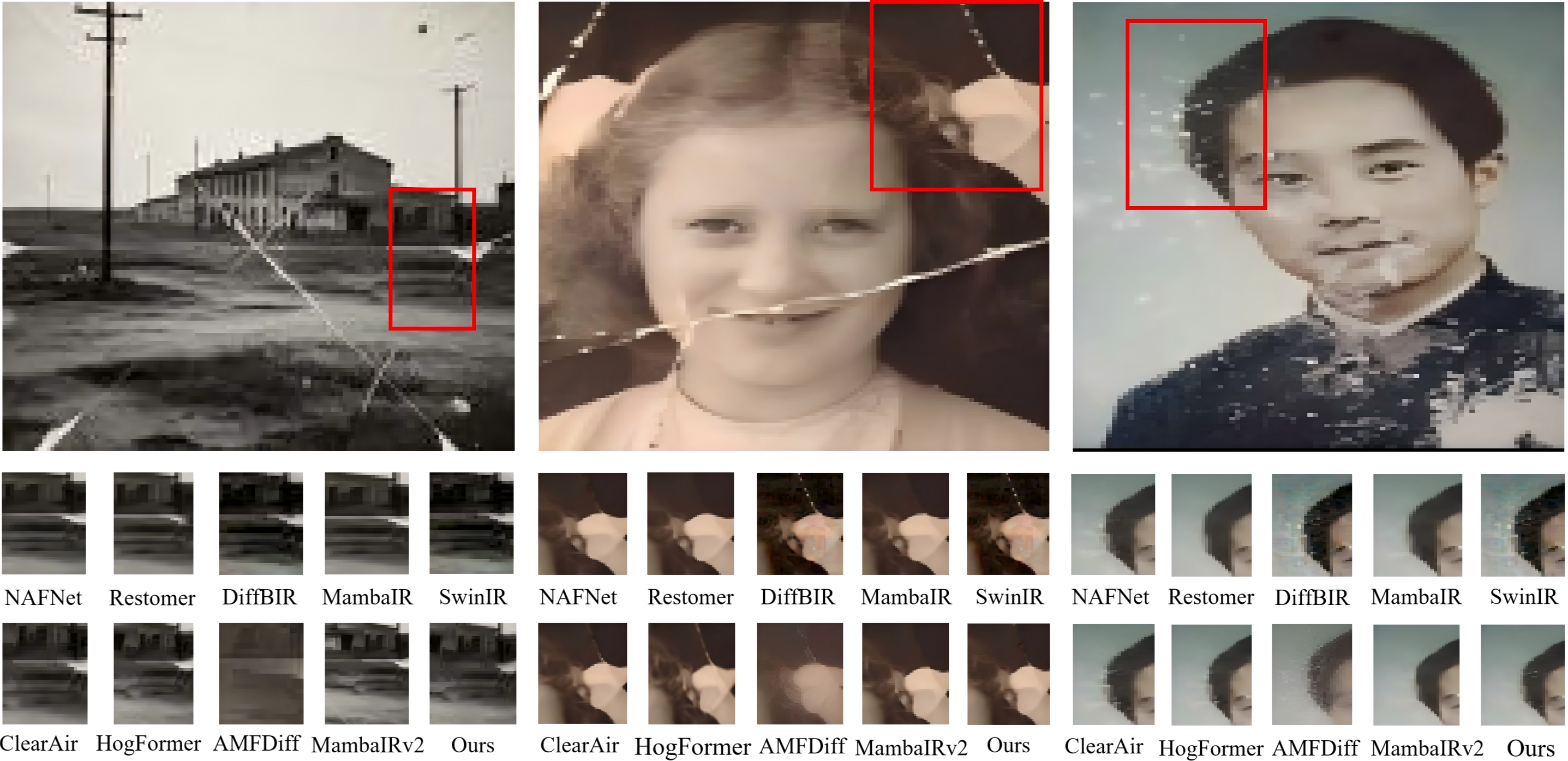}
\caption{Qualitative comparisons on representative real-world old photographs.}
\label{fig5}
\end{figure}

\begin{figure}[t]
\centering
\includegraphics[width=1.0\columnwidth]{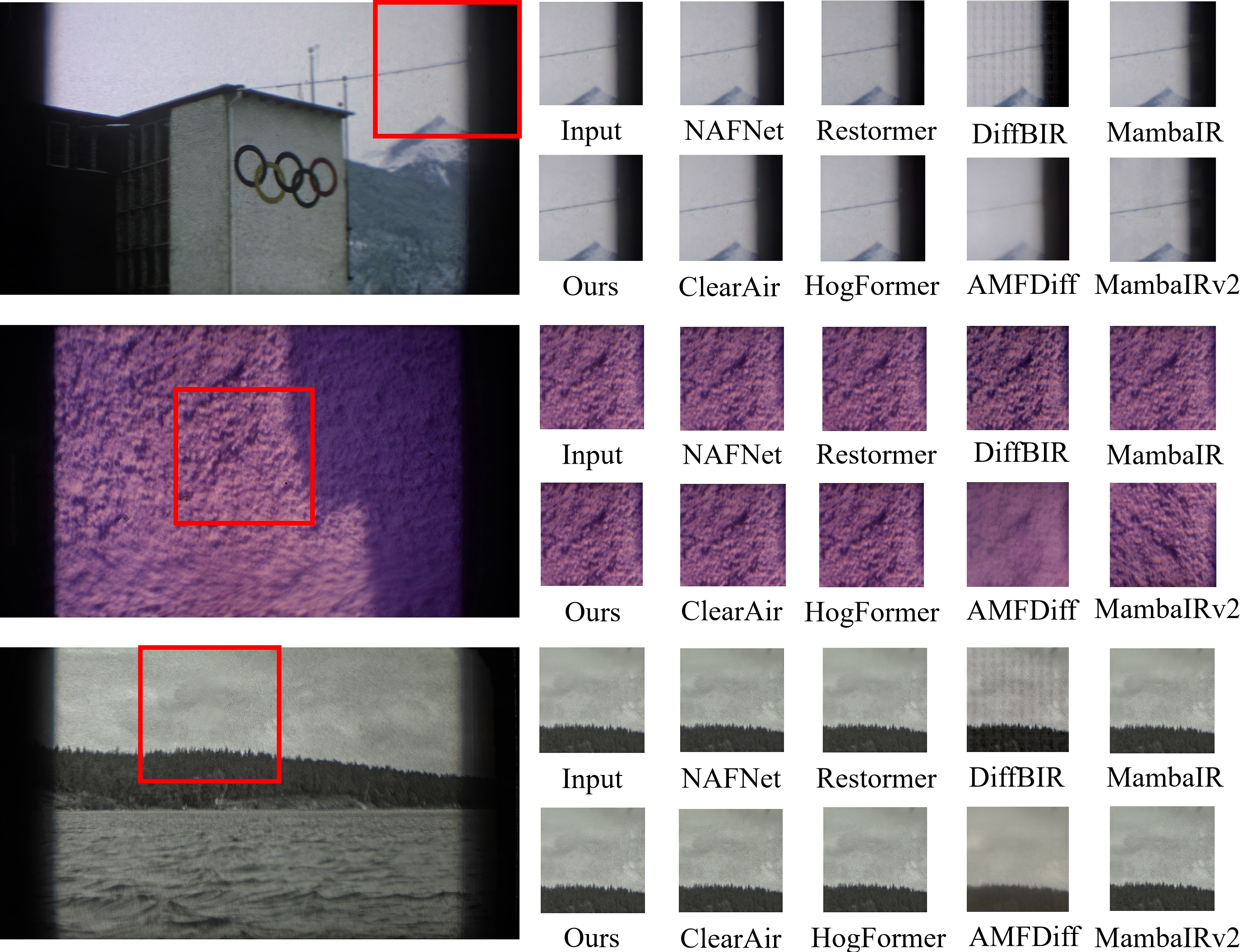}
\caption{Qualitative comparisons on the Stockfilm real-world dataset.}
\label{realworld}
\end{figure}

Beyond quantitative metrics, we provide extensive visual comparisons to intuitively showcase the restoration capabilities. Fig.~\ref{fig5} presents three representative real‑world old photographs—two portrait images and one landscape scene—restored by all competing methods. As can be observed, Loop‑Mamba consistently produces cleaner structural details, richer textures, and more natural color rendition, while effectively suppressing various degradations such as noise, blur, and color shifts. In contrast, other methods either over‑smooth fine details or fail to remove certain artifacts. To further validate generalization, Fig.~\ref{realworld} displays visual results on the Stockfilm dataset. Our Loop‑Mamba again shows superior performance, preserving high‑frequency components and delivering visually pleasing outputs with fewer distortions. 

\begin{table*}[t]
\centering
\small
\setlength{\tabcolsep}{6pt}
\begin{tabular}{l l c c c c}
\toprule
\textbf{Category} & \textbf{Method} & \textbf{Venue \& Year} & \textbf{BRISQUE$\downarrow$} & \textbf{TOPIQ$\uparrow$} & \textbf{MAN-IQA$\uparrow$} \\
\midrule
\multirow{2}{*}{CNN-based} 
  & NAFNet   & ECCV'22  & 27.8305 & 0.3596 & 0.1233 \\
  & ClearAIR & AAAI'26  & 28.6961 & 0.3596 & 0.1204 \\
\midrule
\multirow{2}{*}{Transformer-based} 
  & Restormer & CVPR'22 & 32.5864 & 0.3729 & 0.1388 \\
  & HogFormer & AAAI'26 & 25.1794 & 0.3617 & 0.1246 \\
\midrule
\multirow{2}{*}{Diffusion-based} 
  & DiffBIR  & CVPR'24 & 30.4014 & 0.4211 & 0.1163 \\
  & AMFDiff  & SPIC'26 & 20.9117 & 0.4138 & 0.1969 \\
\midrule
\multirow{3}{*}{Mamba-based} 
  & MambaIR   & ECCV'24 & 30.6522 & 0.3700 & 0.1248 \\
  & MambaIRv2 & CVPR'25 & 30.1474 & 0.3728 & 0.0544 \\
  & \textbf{Loop-Mamba (Ours)} & \textbf{-} & \textbf{22.3574} & 0.3973 & 0.1296 \\
\bottomrule
\end{tabular}
\caption{
Quantitative comparison of no-reference image quality assessment metrics on the Stockfilm real-world dataset.}
\label{tab:quality}
\end{table*}

\subsection{Mobile Deployment}
To further demonstrate the practical applicability of Loop-Mamba, we deploy the proposed model on a Mi 10 Lite Zoom smartphone with an ARM64 processor using ONNX Runtime. As shown in Fig.~\ref{fig7}, Loop-Mamba performs on-device restoration of real degraded photographs while maintaining satisfactory visual quality, demonstrating its feasibility for efficient mobile deployment.
Additional qualitative results, including parameter ablation visualizations and downstream object detection examples, are presented in the supplementary material.

\begin{figure}[H]
\centering
\includegraphics[width=1.0\columnwidth]{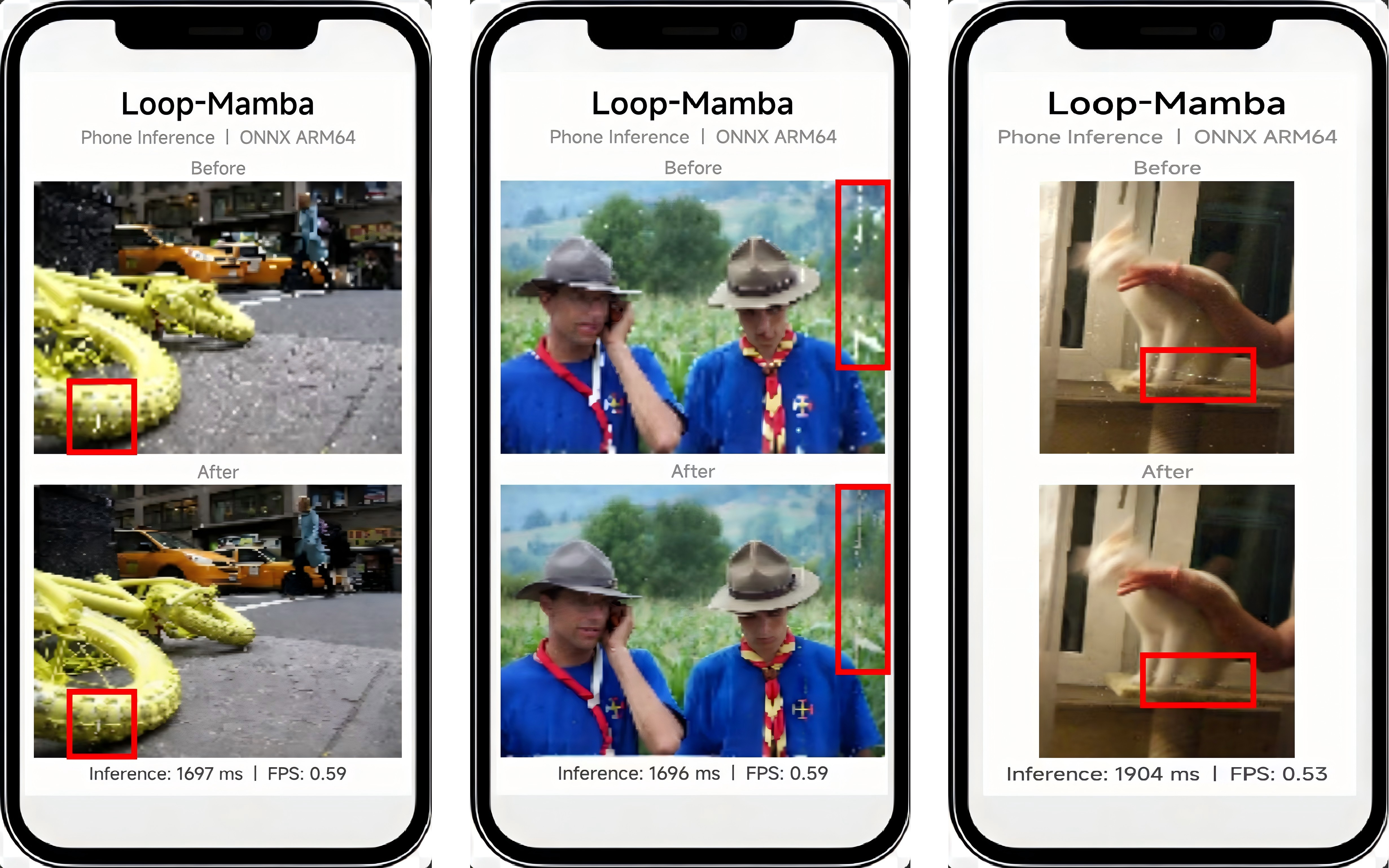}
\caption{Mobile deployment of Loop-Mamba on an ARM64 smartphone using ONNX Runtime.}
\label{fig6}
\end{figure}
\section{Discussion}
\begin{figure}[t]
\centering
\includegraphics[width=1.0\columnwidth]{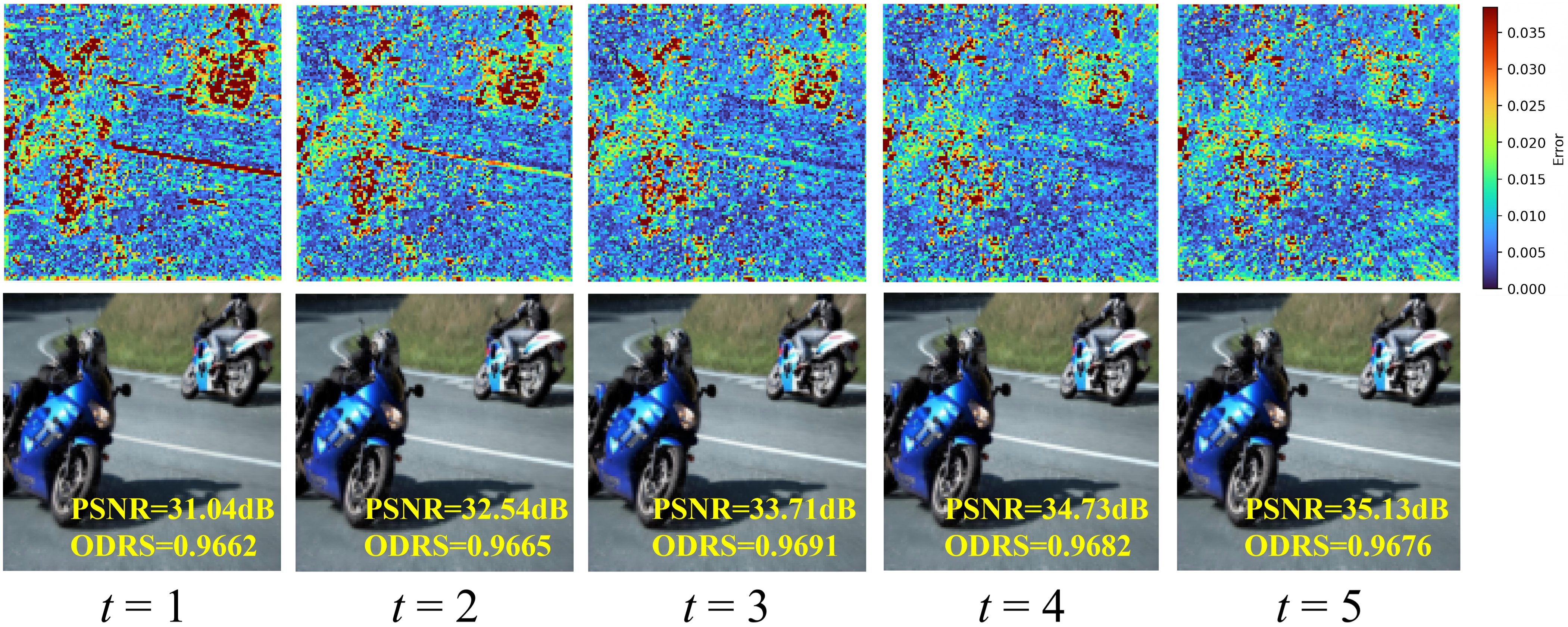}
\caption{Progressive state evolution of Loop-Mamba across five restoration cycles. Error maps show progressively suppressed high-error responses (red/yellow), demonstrating effective error correction through recurrent state propagation.}
\label{fig7}
\end{figure}

To better understand the progressive restoration behavior of Loop-Mamba, Fig.~\ref{fig7} visualizes the restoration results and corresponding error maps across restoration cycles. As the hidden state evolves, the error maps become progressively smoother, indicating that residual errors are gradually eliminated. Meanwhile, both PSNR and ODRS consistently increase, demonstrating effective knowledge accumulation across iterations.
\section{Conclusion}
This paper presented Loop-Mamba, a lightweight state-space framework for old photo restoration based on progressive state evolution. By integrating semantic-guided degradation estimation, shared structural memory, and direction-aware multi-scan propagation, the proposed method effectively models heterogeneous degradations while progressively refining restoration states. We further introduced ODRS, a task-oriented metric that jointly evaluates degradation recovery and structural reconstruction.
\bibliography{aaai2027}
\end{document}